\documentclass[sigplan,screen,authorversion,nonacm]{acmart}
\AtBeginDocument{%
  \providecommand\BibTeX{{%
    \normalfont B\kern-0.5em{\scshape i\kern-0.25em b}\kern-0.8em\TeX}}}

\begin{document}

\title{Aggregated Customer Engagement Model}

\author{Priya Gupta}
\email{priyagu@amazon.com}
\affiliation{%
  \institution{Amazon}
}

\author{Cuize Han}
\email{cuize@amazon.com}
\affiliation{%
  \institution{Amazon}
}

\begin{abstract}
  E-commerce websites use machine learned ranking models to serve shopping results to customers. Typically, the websites log the customer search events, which include the query entered and the resulting engagement with the shopping results, such as clicks and purchases. Each customer search event serves as input training data for the models, and the individual customer engagement serves as a signal for customer preference. So a purchased shopping result, for example, is perceived to be more important than one that is not. However, new or under-impressed products do not have enough customer engagement signals and end up at a disadvantage when being ranked alongside popular products. In this paper, we propose a novel method for data curation that aggregates all customer engagements within a day for the same query to use as input training data. This aggregated customer engagement gives the models a complete picture of the relative importance of shopping results. Training models on this aggregated data leads to less reliance on behavioral features. This helps mitigate the cold start problem and boosted relevant new products to top search results. In this paper, we present the offline and online analysis and results comparing the individual and aggregated customer engagement models trained on e-commerce data. 
\end{abstract}




\maketitle

\section{Introduction}
Retail e-commerce accounts for billions of US dollars in sales worldwide every year. Optimizing product search is critical for customer satisfaction. This means finding the right products to match the customer’s intent, and ranking them in the order that is most important to the customer. Learning to Rank (LTR) \cite{LTR} is a common approach to rank search results. It is a supervised learning algorithm that uses past customer behavior or manual human labels as the signal for product preference.

Product search, unlike web search, poses unique challenges in collecting labeled data. While human annotation of web search results can be useful \cite{eqc}, it can lead to misleading labels in product search. Different product facets might have different importance to customers. For example, price might be a primary factor for some while brand name might be more important to others. The preferences also change based on product type. In studies where human judges were used to rate the products, there was significant disagreement in the ratings and true customer engagement signals  \cite{humanlabels, alonso}.

Using past customer behavior as labels in LTR has similar challenges. For the same query, individual customer engagement (ICE) such as view, click, add-to-cart, add-to-wishlist, purchase etc. might be different. Hence, this ICE based data leads to different labels for the same query-product pairs. With the new model, we propose using aggregated customer engagements (ACE) within a day for the same query across different customer search sessions as labels. The labels now encode the total number of times all customers engaged with a product for the same query within a day. This aggregation gives an estimate of how important a product is in relation to other products since customer engagement is our proxy for relevance. This relative importance is the fundamental idea behind pairwise ranking of LambdaRank algorithm \cite{LambdaRank}.

Customer implicit feedback is known to suffer from position and selection bias \cite{CFR}. Position bias occurs because people are more likely to examine the products that are ranked high. Since customers can only interact with the products that are presented to them, the implicit feedback is biased towards products that are selected by the current ranking model. This bias leads to the “rich get richer and poor get poorer” phenomenon. This is aggravated by the ranking model’s heavy dependence on behavioral features. These are features that directly capture the historic customer behavior data and memorize their preference. While behavioral features help with variance reduction when fitting the ranking model to the customer feedback data, over-reliance on these features leads to less generalizability. For new products or rarely purchased products that do not have enough customer engagement, a model that predominantly ranks based on behavioral features might rank them poorly. In this paper, we show empirically and theoretically, that the ACE model relies less on behavioral features and is better at ranking new or under-impressed products. Long term this can mitigate the impact of position bias.

The rest of the paper is organized as follows: In section ~\ref{method} we describe the methodology behind ACE model. Then we share results comparing the ICE and ACE models in section ~\ref{results}. In section ~\ref{theory}, we dive deep into the theoretical explanation of the efficacy of the ACE model. The final section ~\ref{conclusion} details our conclusions and future work.
\section{ACE Model}
\label{method}

For each customer search event with a query $q_i$, a list of $N_i$ products $\{A_{ij}, j=1,.,N_i\}$ are returned in search results. Each of these products $A_{ij}$ has a feature vector $x_{ij}=(x^1_{ij},..x^d_{ij})$ associated with it that the LTR model uses for ranking. Customers may engage with some products in the search results. This engagement is used as labels $l_{ij}$ for the training data. In case of binary labeling, if a customer engaged with a product, its label $l_{ij} = 1$, otherwise it is zero. In case of the ICE model, each query $q_i$, the corresponding products in search results $\{A_{ij}, j=1,.,N_i\}$ and associated feature vectors $\{x_{ij}, j=1,.,N_i\}$ and labels $\{l_{i1}.,.l_{iN_i}\}$ form an instance of ICE training data. We use millions of instances to train and test the ICE model. 

 In the ACE model, we aggregate the labels for the same query-product pairs $(q,A)$ across different customer search events within the same day:
\[\tilde{l}_{q,A} = \sum_{ (q_i,A_j) = (q,A)} l_{ij}\]
 The motivation behind selecting one day as the aggregation window was to have a large enough window such that the aggregation includes several instances of query-products pairs even for infrequent queries, but small enough window to avoid day-over-day trends in customer preferences in the aggregation. The aggregation of labels can lead to unbounded values for $\tilde{l}_{q,A}$, particularly for popular query-product pairs that occur tens, hundreds or even thousand times in the data. So we bucket these aggregated labels $\tilde{l}_{q,A}$ using quantiles and map onto a finite set of integer labels. For ACE model, each unique daily query along with all the corresponding distinct search results and their associated features and the aggregated labels form an instance of ACE training data. Like the ICE model, we use millions of ACE instances to train and test the ACE model.

The capping of labels in ACE data has the consequence of limiting the influence of highly popular query-product pairs on the ranking model. In other words, down-sampling the signals from these frequently impressed query-product pairs leads to a ranking model that relies less on behavioral features. As a result, products that have little engagement, either because they are new or preferred by a minority of customers, have a chance to move to top search results. This is how the ACE model helps alleviate the cold start problem that is common in product search.

\section{Results}
\label{results}

We trained ICE and ACE models on three data sets from a large e-commerce website. Each set had more than 3M data points. We tested the models offline and ran A/B tests online. With each experimental result, we iterated over the ACE method to improve its efficacy at mitigating the cold start problem without impacting other customer engagement related metrics.

Aggregating customer engagement and collapsing several searches with the same query into one data point leads to a larger percentage of distinct queries and products in the data. For example, in one of our data sets 86\% of the queries in the ACE data were distinct while only 49\% in the ICE data were unique. Similarly, 89\% of the query-product pairs in the ACE data were distinct while 62\% in ICE data were unique. Greater diversity in the data leads to smaller errors and more generalized models \cite{diversity}. 

During model assessment, we measure the variance reduction in the labels due to each feature selected by the model. This variance reduction is correlated with feature importance. We compared the total variance reduction due to all the behavioral features in the ACE and ICE models trained on the three data sets. The larger the variance reduction due to a feature, the more the model relies on it for ranking. As is clear in Table. ~\ref{tab:variance}, the ACE models rely less on behavioral features when compared to ICE model. Although the difference is small, it is enough to allow more textual and product newness (days since launch date) related features to be picked up by the ACE model. This helps surface products that are better exact matches with the query even if they are new and have not accumulated enough customer engagement related behavioral features. 
\begin{table} [h]
  \caption{Total variance reduction on model prediction due to all behavioral features picked up by feature selection during model development}
  \label{tab:variance}
  \begin{tabular}{ccl}
    \toprule
    Dataset&ICE Model&ACE Model\\
     \midrule
    1 & 85.34\% &\hfil  80.54\%\\
    2&76.8\% &\hfil  76.38\%\\
    3 & 83.23\% &\hfil  80.81\%\\
 
  \bottomrule
\end{tabular}
\end{table}

In the offline analysis of the models, we measured how effective the ACE model was at surfacing newer products, that were less than seven days old, in top 16 search results. When compared to the ICE model, on one data set the ACE model served 30\% more new product impressions while on another data set it served 110\% more. The big jump from 30\% to 110\% came due to an improvement in feature selection strategy that was applied to the latter data set. 

After incorporating improvements and learnings from offline analysis, we ran an A/B test on the e-commerce website in the US. We measured the difference between new product impressions, clicks and purchases between the ICE and ACE models. For this online experiment, 'new product' was defined as any product launched on the e-commerce website in the past three days. The experiment ran for two weeks during which the ACE model served 12.4\% more new product impressions than the ICE model when measured over millions of customer search sessions. Because of the increase in new product impressions, the ACE model led to 10.6\% more clicks and 17.54\% more number of purchases of these new products. These results validated our hypothesis that the ACE model can help with the cold start problem.

\section{Theoretical Explanation}
\label{theory}

The ACE model down-samples the signals from highly impressed products whose behavioral features have enough power to predict the customer action well. Thus, models trained on ACE data focus more on non-behavioral features which
leads to better generalizability.

Given a query $q$, denote $A$ to be a random selected product that
presented to the customer. Let $Y$ to be the $0/1$ random variable
that represents the customer action for that pair. The feature vector
of the pair can be transformed and decomposed into behavioral features ($X_{bhv}$) and non-behavioral features ($X_{nbhv}$) that are random elements in $\mathcal{X}$. For $(x,x')\in\mathcal{X}$,
denote the conditional expectations as

\begin{align*}
f_{bhv}(x)= & \mathbb{E}\left[Y|X_{bhv}=x\right]\\
f_{nbhv}(x')= & \mathbb{E}\left[Y|X_{nbhv}=x'\right]
\end{align*}
Then $f_{bhv}(X_{bhv})$ and $f_{nbhv}(X_{nbhv})$ are theoretically
the best prediction of the customer action given the knowledge of
$X_{bhv}$ or $X_{nbhv}$. We also assume the effects have been orthogonalized
and the residues are uncorrelated: 
\begin{equation}
\mathbb{E}\left(Y-\mathbb{E}\left[Y|X_{bhv}\right]\right)\left(Y-\mathbb{E}\left[Y|X_{nbhv}\right]\right)=0\label{eq: uncorrelation assumption}
\end{equation}
Suppose the scoring function behind the ranking takes
the form of $f_{w}(x,x')=wf_{bhv}(x)+(1-w)f_{nbhv}(x')$ with a weight
parameter $w\in[0,1]$, then the optimal weight that minimizes the
expected mean square error can be represented as :
\begin{align}
\mathbb{E}\left[\left(Y-f_{w}(X_{bhv},X_{nbhv})\right)^{2}\right] 
 & =w^{2}\mathbb{E}\epsilon_{bhv}(X_{bhv})+ \nonumber \\
 & (1-w)^{2}\mathbb{E}\epsilon_{nbhv}(X_{nbhv})\label{eq: variance decompose}
\end{align}
where $\mathbb{E}\epsilon_{bhv}(X_{bhv})$ and $\mathbb{E}\epsilon_{nbhv}(X_{nbhv})$
represent the variances that cannot be explained by only $X_{bhv}$
or $X_{nbhv}$ respectively and 
\begin{align*}
\epsilon_{bhv}(x)=&\mathbb{E}\left[(Y-f_{bhv}(X_{bhv}))^{2}|X_{bhv}=x\right]\\
\epsilon_{nbhv}(x')=&\mathbb{E}\left[(Y-f_{nbhv}(X_{nbhv}))^{2}|X_{nbhv}=x'\right].
\end{align*}
 (\ref{eq: variance decompose}) also uses the assumption (\ref{eq: uncorrelation assumption}).
Then solving the quadratic function of $w$ in (\ref{eq: variance decompose}),
we have the optimal weight $w^{*}$ equals
\begin{equation}
w^{*}=\dfrac{\mathbb{E}\epsilon_{nbhv}(X_{nbhv})}{\mathbb{E}\epsilon_{bhv}(X_{bhv})+\mathbb{E}\epsilon_{nbhv}(X_{nbhv})}\label{eq: optimal weight}
\end{equation}
So the model will put less weight on behavioral
features if non-behavioral features get better ability to reduce
the variance (corresponding to the decrease of $\mathbb{E}\epsilon_{nbhv}(X_{nbhv})$
) or/and behavioral features reduce less variance (corresponding to
the increase of $\mathbb{E}\epsilon_{bhv}(X_{bhv})$).

We claim that for the current data distribution,
ACE will decrease $\mathbb{E}\epsilon_{nbhv}(X_{nbhv})$
and increase $\mathbb{E}\epsilon_{bhv}(X_{bhv})$. Consequently, the model will put more weight on non-behavioral features.

Suppose there are $m$ customer search sessions with the query $q$ in the training data. Given $m$ labels $Y_{1},..,Y_{m}$, ACE model makes a new label
$\tilde{Y}$ out of $Y_{1},..,Y_{m}$ for the product $A$ and features
$(X_{bhv},X_{nbhv})$. For simplicity, we consider the extreme cut-off
situation where we define 
\begin{equation}
\tilde{Y}=\begin{cases}
1 & \text{at least one of \ensuremath{Y_{1},..,Y_{m}} is \ensuremath{1}}\\
0 & \text{otherwise}
\end{cases}\label{eq: extreme cutoff label}
\end{equation}
 Then for a feature $X$ if we denote $p(x)=\mathbb{P}(Y=1|X=x),$ then for the new label 
\begin{align}
\tilde{p}(x) & =\mathbb{P}(\tilde{Y}=1|X=x)\nonumber \\
 & =1-\mathbb{P}(\tilde{Y}=0|X=x)\nonumber \\
 & =1-\left(\mathbb{P}(Y=0|X=x)\right)^{m}\nonumber \\
 & =1-(1-p(x))^{m}\label{eq: new label prob}
\end{align}
Now the variance that cannot be explained by $X$ for label $Y$ is $\mathbb{E}\epsilon(X)$
where 
\begin{align}
\epsilon(x) & =\mathbb{E}\left[\left(Y-\mathbb{E}(Y|X)\right)^{2}|X=x\right]\nonumber \\
 & =\mathbb{E}[Y^{2}|X=x]-(\mathbb{E}(Y|X=x))^{2}\nonumber \\
 & =\mathbb{E}[Y|X=x]-(\mathbb{E}(Y|X=x))^{2}\nonumber \\
 & =p(x)(1-p(x))\label{eq: epsilon x and p(x)}
\end{align}
The derivation of (\ref{eq: epsilon x and p(x)}) uses the fact that
$Y^{2}\equiv Y$ and $\mathbb{E}[Y|X=x]\equiv\mathbb{P}(Y=1|X=x)$.
Then the corresponding quantity for $\tilde{Y}$ is 
\begin{align}
\tilde{\epsilon}(x) & =\mathbb{E}\left[\left(\tilde{Y}-\mathbb{E}(\tilde{Y}|X)\right)^{2}|X=x\right]\nonumber \\
 & =\tilde{p}(x)(1-\tilde{p}(x))\nonumber \\
 & =(1-(1-p(x))^{m})(1-p(x))^{m}\label{eq: e tilde x with p(x)}
\end{align}
\begin{figure}
\caption{Plot of unexplained variance in ACE model $\tilde{\epsilon}(x)$ against probability of customer action $p(x)$ when aggregating different number of customer sessions m. This figure shows that as number of sessions aggregated goes up, the unexplained variance quickly drops to zero as probability of customer action increases.}
\includegraphics[width=\linewidth]{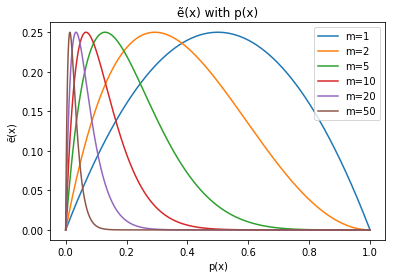}
\label{fig:epplot}
\end{figure}
The error $\epsilon(x)$ will be small if $p(x)$ is close to $0$ or $1$, that is, given $X=x$, we are able to tell
with more certainty whether the customer will take action or
not. On the other hand, it is maximized when $p(x)=1/2$, corresponding
to given $X=x$, we are still randomly guessing the result. So the
plot between $\epsilon(x)$ and $p(x)$ looks like a bell curve and
with the effect of ACE, the curve skews to the left with the increase
of aggregated session number $m$ as shown in Figure \ref{fig:epplot}.

We observe that
currently majority of the variance of the label is explained by the
behavioral features. That is $\mathbb{E}\epsilon_{bhv}(X_{bhv})$,
the unexplained variance of behavioral features is much smaller than
$\mathbb{E}\epsilon_{nbhv}(X_{nbhv})$. In order for $\mathbb{E}\epsilon_{bhv}(X_{bhv})$ to be
small, the majority of the mass of the distribution of $X_{bhv}$
needs to accumulate at the two ends of $p(x)$. Since the label
$Y$ itself skews to the left (most products won't receive action), we
claim that for behavioral features, majority of the mass of the distribution
accumulates to the left end of $p(x)$. From Figure \ref{fig:epplot}, we see that
the error $\tilde{\epsilon}(x)$ increases at the left end of $p(x)$
with the increase of $m$. Thus, the unexplained
variance for behavioral features in ACE becomes larger: $\mathbb{E}\tilde{\epsilon}_{bhv}(X_{bhv})>\mathbb{E}\epsilon_{bhv}(X_{bhv})$.
On the other hand, relatively most of the mass of $X_{nbhv}$ concentrates
in the middle part of $p(x)$ as the unexplained variance of $X_{nbhv}$
is large. From the figure we see at the middle part of $p(x)$, the
error $\tilde{\epsilon}(x)$ quickly drop to near zero as $m$ increases.
Thus, for non-behavioral features, the unexplained variance
becomes smaller: $\mathbb{E}\tilde{\epsilon}_{nbhv}(X_{nbhv})<\mathbb{E}\epsilon_{nbhv}(X_{nbhv})$.
Hence, the optimal weight $\tilde{w^{*}}$ on the behavioral features
in ACE reduces as 
\[
\dfrac{\mathbb{E}\tilde{\epsilon}_{nbhv}(X_{nbhv})}{\mathbb{E}\tilde{\epsilon}_{bhv}(X_{bhv})+\mathbb{E}\tilde{\epsilon}_{nbhv}(X_{nbhv})}<\dfrac{\mathbb{E}\epsilon_{nbhv}(X_{nbhv})}{\mathbb{E}\epsilon_{bhv}(X_{bhv})+\mathbb{E}\epsilon_{nbhv}(X_{nbhv})}
\]

\section{Conclusion}
\label{conclusion}

In this paper, we described a new way of processing training data for developing ranking models for e-commerce. We showed that aggregating customer engagement across different search sessions for the same query leads to better ranking of new and under-impressed products. This is because the ACE model relies less on behavioral features which otherwise tend to dominate the ranking. Products that are popular have the benefit of position and selection bias leading to strong behavioral features. With a model that under-samples these customer engagement signals, we can mitigate the cold start problem. Products that are under-impressed, perhaps because they are preferred by a minority of customers, also get a chance to show in top search results. We plan to conduct more online experiments with the ACE model to further validate the methodology using empirical results.

\bibliographystyle{ACM-Reference-Format}
\bibliography{ACE}


\begin{thebibliography}{7}


\ifx \showCODEN    \undefined \def \showCODEN     #1{\unskip}     \fi
\ifx \showDOI      \undefined \def \showDOI       #1{#1}\fi
\ifx \showISBNx    \undefined \def \showISBNx     #1{\unskip}     \fi
\ifx \showISBNxiii \undefined \def \showISBNxiii  #1{\unskip}     \fi
\ifx \showISSN     \undefined \def \showISSN      #1{\unskip}     \fi
\ifx \showLCCN     \undefined \def \showLCCN      #1{\unskip}     \fi
\ifx \shownote     \undefined \def \shownote      #1{#1}          \fi
\ifx \showarticletitle \undefined \def \showarticletitle #1{#1}   \fi
\ifx \showURL      \undefined \def \showURL       {\relax}        \fi
\providecommand\bibfield[2]{#2}
\providecommand\bibinfo[2]{#2}
\providecommand\natexlab[1]{#1}
\providecommand\showeprint[2][]{arXiv:#2}

\bibitem[\protect\citeauthoryear{Aman~Agarwal}{Aman~Agarwal}{2019}]%
        {CFR}
\bibfield{author}{\bibinfo{person}{Ivan Zaitsev Thorsten~Joachims Aman~Agarwal,
  Kenta~Takatsu}.} \bibinfo{year}{2019}\natexlab{}.
\newblock \showarticletitle{A General Framework for Counterfactual
  Learning-to-Rank}.
\newblock \bibinfo{journal}{\emph{In Proceedings of the 42nd International ACM
  SIGIR Conference on Research and Development in Information Retrieval}}
  (\bibinfo{date}{july} \bibinfo{year}{2019}), \bibinfo{pages}{5--14}.
\newblock


\bibitem[\protect\citeauthoryear{Burges}{Burges}{[n.d.]}]%
        {LambdaRank}
\bibfield{author}{\bibinfo{person}{Christopher~J.C. Burges}.}
  \bibinfo{year}{[n.d.]}\natexlab{}.
\newblock \showarticletitle{From RankNet to LambdaRank to LambdaMART: An
  Overview}.
\newblock \bibinfo{journal}{\emph{Microsoft Research Technical Report}},
  Article \bibinfo{articleno}{MSR-TR-2010-82} (\bibinfo{year}{[n.\,d.]}).
\newblock


\bibitem[\protect\citeauthoryear{John~Le}{John~Le}{2010}]%
        {eqc}
\bibfield{author}{\bibinfo{person}{Vaughn Hester Lukas~Biewald John~Le,
  Andrew~Edmonds}.} \bibinfo{year}{2010}\natexlab{}.
\newblock \showarticletitle{Ensuring quality in crowdsourced search relevance
  evaluation: The effects of training question distribution}.
\newblock \bibinfo{journal}{\emph{Proceedings of the SIGIR 2010 Workshop on
  Crowdsourcing for Search Evaluation}} (\bibinfo{date}{july}
  \bibinfo{year}{2010}).
\newblock


\bibitem[\protect\citeauthoryear{Liu}{Liu}{2011}]%
        {LTR}
\bibfield{author}{\bibinfo{person}{Tie-Yan Liu}.}
  \bibinfo{year}{2011}\natexlab{}.
\newblock \bibinfo{booktitle}{\emph{Learning to Rank for Information
  Retrieval}}.
\newblock \bibinfo{publisher}{Springer}, \bibinfo{address}{New York, NY}.
\newblock


\bibitem[\protect\citeauthoryear{Omar~Alonso}{Omar~Alonso}{2009}]%
        {alonso}
\bibfield{author}{\bibinfo{person}{Stefano~Mizzaro Omar~Alonso}.}
  \bibinfo{year}{2009}\natexlab{}.
\newblock \showarticletitle{Relevance criteria for e-commerce: a
  crowdsourcing-based experimental analysis.}
\newblock \bibinfo{journal}{\emph{Proceedings of the 32nd international ACM
  SIGIR conference on Research and development in information retrieval}}
  (\bibinfo{year}{2009}).
\newblock


\bibitem[\protect\citeauthoryear{Shubhra Kanti Karmaker~Santu}{Shubhra Kanti
  Karmaker~Santu}{2017}]%
        {humanlabels}
\bibfield{author}{\bibinfo{person}{ChengXiang~Zhai Shubhra Kanti
  Karmaker~Santu, Parikshit~Sondhi}.} \bibinfo{year}{2017}\natexlab{}.
\newblock \showarticletitle{On Application of Learning to Rank for E-Commerce
  Search}.
\newblock \bibinfo{journal}{\emph{SIGIR '17: Proceedings of the 40th
  International ACM SIGIR Conference on Research and Development in Information
  Retrieval}} (\bibinfo{date}{aug} \bibinfo{year}{2017}),
  \bibinfo{pages}{475--484}.
\newblock


\bibitem[\protect\citeauthoryear{Trevor~Hastie}{Trevor~Hastie}{2009}]%
        {diversity}
\bibfield{author}{\bibinfo{person}{Jerome~Friedman Trevor~Hastie,
  Robert~Tibshirani}.} \bibinfo{year}{2009}\natexlab{}.
\newblock \bibinfo{booktitle}{\emph{The Elements of Statistical Learning}
  (\bibinfo{edition}{2nd} ed.)}.
\newblock \bibinfo{publisher}{Springer}, \bibinfo{address}{New York, NY}.
\newblock


\end{thebibliography}

\end{document}